\documentclass[pdflatex,sn-nature]{sn-jnl}
\usepackage{graphicx}
\usepackage{amsmath,amssymb}
\usepackage{booktabs}
\usepackage{url}
\graphicspath{{./}}
\setcitestyle{super,open={},close={},sort&compress}

\title[Physics-Guided PredHydro-Net]{Physics-Guided Dual Decoding and Spectral Supervision for Global 3D Hydrometeor Prediction}

\author[1,2]{\fnm{Dandan} \sur{Chen}}

\author*[1,2]{\fnm{Yaqiang} \sur{Wang}}\email{yqwang@cma.gov.cn}

\affil[1]{\orgdiv{State Key Laboratory of Severe Weather Meteorological Science and Technology}, \orgname{Chinese Academy of Meteorological Sciences}, \orgaddress{\city{Beijing}, \country{China}}}

\affil[2]{\orgdiv{Xiong'an Institute of Meteorological Artificial Intelligence}, \orgaddress{\city{Xiong'an}, \country{China}}}

\begin{document}

\abstract{While global data-driven models excel at predicting continuous atmospheric variables, three-dimensional hydrometeor forecasting remains challenging due to the zero-inflated, long-tailed distributions of these variables. Standard deep learning optimization often yields overly smooth forecasts, attenuating extreme events and spatial textures. We propose PredHydro-Net, a physics-guided dual-decoding framework that mitigates this smoothing. To resolve multi-variable optimization conflicts, it employs a decoupled architecture where macroscopic thermodynamic and dynamic fields unidirectionally modulate hydrometeor generation. By integrating wavelet-based frequency decoupling, spectral amplitude matching, and adversarial training, the model achieves a favorable trade-off between quantitative accuracy and spatial fidelity. In a 72-h global evaluation, PredHydro-Net outperforms both spatiotemporal deep learning baselines (Earthformer and PredRNNv2) and the operational Global Forecast System (GFS) in extreme-event detection and spectral representation. Furthermore, it demonstrates strong climatological consistency with Global Precipitation Measurement (GPM) satellite retrievals. The model reasonably reproduces the three-dimensional cloud structures in extreme weather events, such as Hurricane Ian. Feature attribution confirms its dependence on physical precursors such as relative humidity and wind convergence, offering a robust, physics-informed approach to long-tailed atmospheric prediction.}


\maketitle

\section{Introduction}

The three-dimensional distribution and evolution of atmospheric hydrometeors, including cloud ice, cloud liquid water, rain, and snow, are central to Earth's radiative balance, large-scale circulation, and extreme weather events.\cite{ceppi2017cloud,matus2017role,cossu2014influence,morrison2020confronting} In numerical weather prediction (NWP), accurate hydrometeor representation directly affects forecasts of convection, precipitation intensity, and aviation weather hazards.\cite{liu2007sensitivity,van2012quantification,karki2018wrf,mctaggart2019modernization} However, traditional NWP systems represent these processes through complex cloud microphysics parameterizations, which are computationally expensive and remain uncertain at subgrid scales.\cite{hong2012next,morrison2015parameterization,field2023implementation,brotzge2023challenges}

In recent years, data-driven weather prediction has developed rapidly. Global weather models such as Pangu-Weather,\cite{bi2023accurate} GraphCast,\cite{lam2023learning} and FourCastNet\cite{pathak2022fourcastnet,bonev2025fourcastnet} have shown the potential to approach, and in some metrics outperform, operational dynamical models for continuous atmospheric variables such as temperature ($T$), geopotential height ($H$), and horizontal winds ($u/v$).\cite{chen2023fuxi,price2025probabilistic} However, global three-dimensional prediction of hydrometeor variables remains comparatively less explored.

Hydrometeor prediction poses a more complex optimization problem than conventional temperature and pressure fields. Hydrometeors are strongly zero-inflated and non-Gaussian: most clear-sky regions contain near-zero hydrometeor content, whereas extreme values are concentrated in relatively narrow convective bands.\cite{marchand2012spatial,zhang2018multiband,barlakas2021introducing,hess2022deep} This data property can lead to a multi-variable optimization dilemma. In a shared-decoder architecture, standard mean squared error (MSE) optimization may be dominated by dense continuous variables, such as temperature and wind, causing zero-inflated hydrometeors to regress toward the mean and lose high-frequency texture.\cite{weyn2021sub,chen2022rainnet,iotti2025rainscalegan,leinonen2020stochastic,glawion2025global,hess2025fast} Conversely, excessively upweighting sparse extreme hydrometeor values may introduce noisy high-frequency gradients into the shared latent space, potentially causing negative transfer and degrading other continuous variables. This optimization conflict creates an inherent trade-off, making it difficult for a single global model to concurrently resolve smooth background fields and extreme hydrometeor structures.

To mitigate this problem, we propose PredHydro-Net, a physics-guided dual decoding model for global hydrometeor prediction. The main contributions are threefold.

First, we introduce a physics-guided decoupling strategy. Rather than using a fully shared black-box decoder for all variables, we design a feature-wise linear modulation (FiLM) module, termed TQ2HydroFiLM, that uses predicted continuous thermodynamic fields, including temperature and moisture fields, as physical boundary conditions to unidirectionally modulate the hydrometeor branch. This design is intended to represent the constraint imposed by the macroscopic background state on cloud and precipitation formation, improving consistency with basic physical relationships.

Second, we introduce multi-scale spectral supervision. A Haar discrete wavelet transform (DWT) decomposes hydrometeor fields into low-frequency envelopes and high-frequency texture components. By combining a spectral amplitude loss with a patch-based generative adversarial network (PatchGAN) loss, we aim to improve forecast accuracy while better preserving high-frequency spectral characteristics in hydrometeor fields.

Third, we provide physical interpretability for deep learning-based cloud and precipitation generation. Through global three-dimensional prediction experiments and gradient-based attribution, we explicitly validate the driving role of macroscopic physical features, such as relative humidity (RH) and wind convergence, confirming that the model learns physically consistent pathways rather than purely statistical correlations.

The results show that in a global 72-h evaluation, PredHydro-Net improves selected hydrometeor error and extreme-event metrics relative to the operational Global Forecast System (GFS) baseline, and it shows advantages in representing three-dimensional hydrometeor structures in selected extreme weather cases, including hurricanes and Dragon-Boat rainfall events. This study provides a direction for further investigation of long-tailed and spatially discontinuous atmospheric variables in global data-driven models.

\section{Results}

\subsection{PredHydro-Net Framework}

\begin{figure}[!htbp]
\centering
\includegraphics[width=\linewidth]{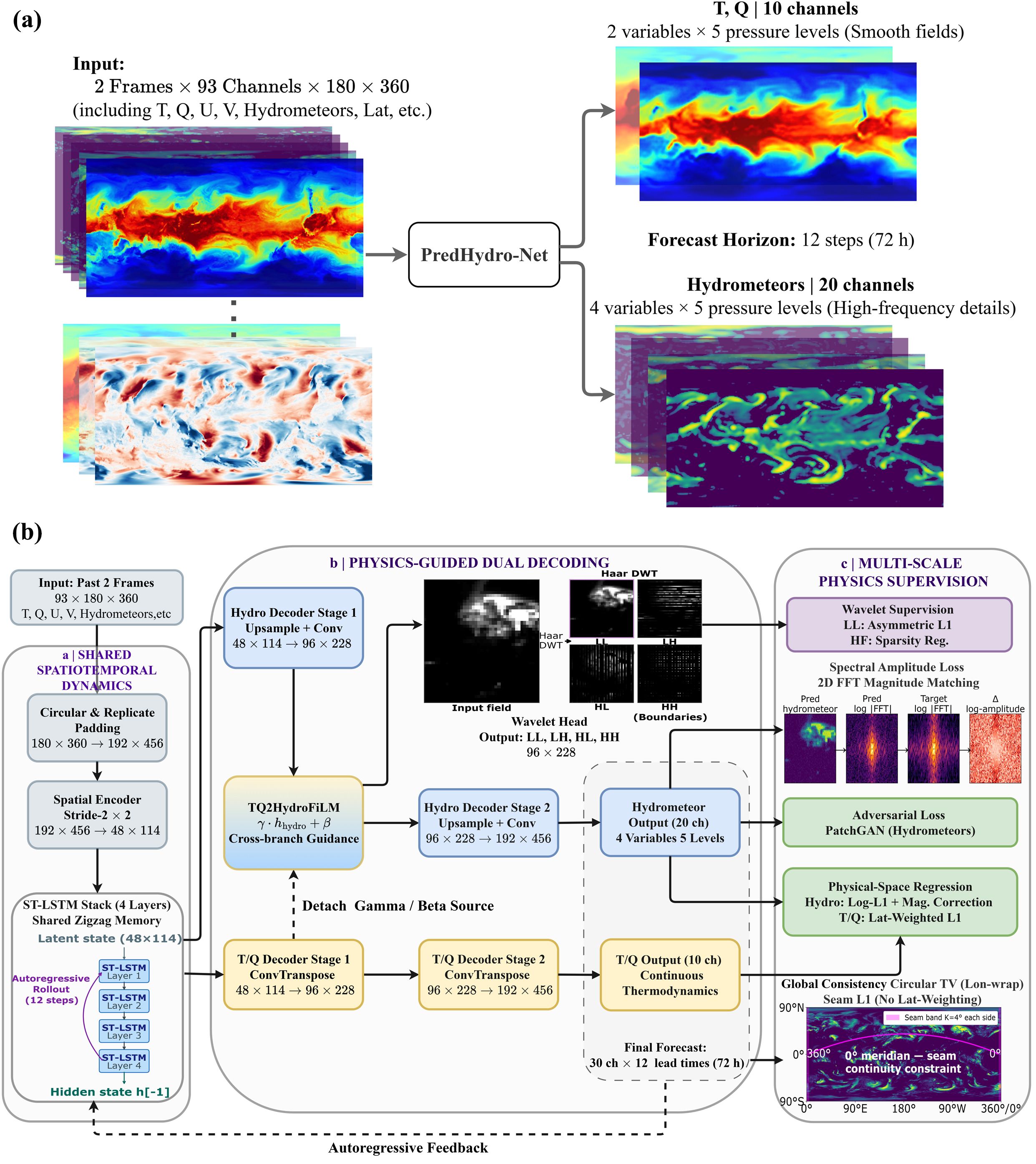}
\caption{Overall architecture of PredHydro-Net. \textbf{a}, Macroscopic sequence-to-sequence prediction framework, mapping two input atmospheric states with 93 channels to future thermodynamic and hydrometeor states over a 72-h horizon. \textbf{b}, Schematic of the physics-guided dual decoding module. Shared spatiotemporal dynamics are extracted by an ST-LSTM stack, while the thermodynamic branch provides detached cross-branch guidance to hydrometeor generation through the TQ2HydroFiLM module. A Haar discrete wavelet transform (DWT) head provides auxiliary low- and high-frequency hydrometeor supervision, while the final hydrometeor output is produced by learned upsampling and convolution. Additional supervision includes spectral amplitude matching, pixel-space PatchGAN adversarial training on hydrometeor channels, physical-space regression, and a 0$^{\circ}$/360$^{\circ}$ seam-consistency constraint.}
\label{fig:architecture}
\end{figure}

PredHydro-Net is built on the PredRNNv2\cite{wang2022predrnn} spatiotemporal prediction architecture (Fig.~\ref{fig:architecture}). To capture atmospheric evolution, the model uses a four-layer spatiotemporal long short-term memory (ST-LSTM) stack to extract hidden-state representations. The input consists of two past time steps ($T_{\rm in}=2$) with 93 channels, including 92 European Centre for Medium-Range Weather Forecasts Reanalysis v5 (ERA5)-derived meteorological channels and one appended latitude channel. The output consists of twelve future time steps ($T_{\rm out}=12$), corresponding to 72 h, with 30 target channels: 10 continuous thermodynamic channels and 20 hydrometeor channels. The hydrometeor targets cover four species: specific cloud ice water content (CIWC), specific cloud liquid water content (CLWC), specific rain water content (CRWC), and specific snow water content (CSWC), on five pressure levels: 200, 500, 700, 850, and 1000 hPa. The architecture separates dense thermodynamic reconstruction from sparse hydrometeor generation, while allowing the thermodynamic branch to provide one-way physical modulation for hydrometeor prediction.

We evaluate PredHydro-Net for global three-dimensional hydrometeor prediction over a 72-h forecast horizon with 6-h temporal resolution. The model is compared against data-driven baselines (Earthformer\cite{gao2022earthformer} and PredRNNv2\cite{wang2022predrnn}) and the National Centers for Environmental Prediction (NCEP) GFS operational forecast. The evaluation covers statistical errors, extreme-event detection, spatial spectral characteristics, three-dimensional weather-system structure, and global climatological consistency.

\subsection{Quantitative Performance against Operational NWP}

\begin{figure}[!htbp]
\centering
\includegraphics[width=\linewidth]{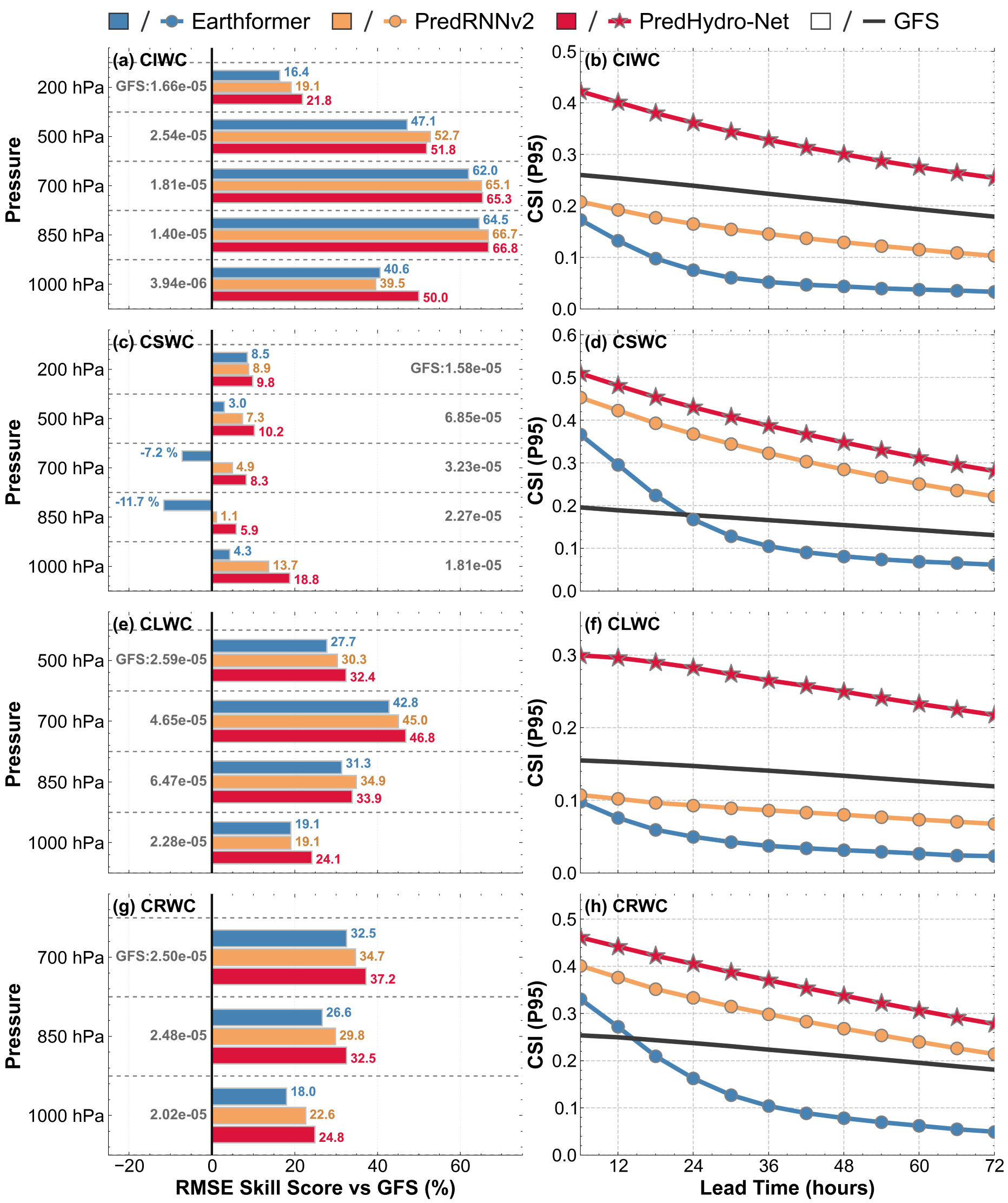}
\caption{Quantitative evaluation of hydrometeor predictions against the GFS baseline. \textbf{a, c, e, g}, Vertical profiles of the root mean square error (RMSE) skill score (\%) of PredHydro-Net, Earthformer, and PredRNNv2 relative to the uniformly processed GFS forecasts for CIWC, CSWC, CLWC, and CRWC at a 72-h lead time. Positive values indicate lower RMSE than GFS, and gray annotations give the corresponding GFS RMSE values in kg kg$^{-1}$. \textbf{b, d, f, h}, Critical success index (CSI) at the 95th percentile threshold as a function of lead time up to 72 h. Paired bootstrap 95\% confidence intervals are omitted because they are narrow relative to the plotted ranges.}
\label{fig:skill_csi}
\end{figure}

For large-scale gridpoint error metrics, deep learning models may benefit from smoothing effects. However, the sparsity of hydrometeor variables complicates this expectation. Figure~\ref{fig:skill_csi}a, c, e, g shows root mean square error (RMSE) skill scores relative to the converted GFS reference for hydrometeor specific content variables (CIWC, CSWC, CLWC, and CRWC) at the evaluated pressure levels over the 72-h forecast horizon.\cite{han2017updates,wallace2006atmospheric} Because some upper-level liquid hydrometeor channels are extremely sparse, the vertical profiles focus on physically meaningful levels for each species. For example, CRWC is reported only from 700 to 1000 hPa. In the 72-h forecast evaluation, PredHydro-Net shows positive skill scores across multiple hydrometeor phases and levels, with the clearest gains appearing for highly skewed hydrometeor variables rather than for all atmospheric fields.

The critical success index (CSI) at the 95th percentile threshold (P95) provides a more demanding test for extreme hydrometeor events (Fig.~\ref{fig:skill_csi}b, d, f, h). The two conventional spatiotemporal deep learning baselines do not consistently outperform GFS in this threshold-based evaluation. Earthformer shows a rapid decline in CSI with lead time, especially for CIWC, CSWC, and CRWC, while PredRNNv2 is more competitive but still remains below PredHydro-Net across the evaluated hydrometeor species. This pattern highlights a limitation of standard MSE-oriented optimization: although such models can reduce average errors by producing smoother fields, their ability to capture intense, localized hydrometeor cores is not guaranteed.

GFS, grounded in explicit physical parameterizations, therefore remains a robust and strong baseline for high-threshold precipitation and cloud systems. In this context, the performance of PredHydro-Net is particularly notable. It not only achieves competitive or improved RMSE relative to the deep learning baselines, but also maintains higher and more stable CSI curves over the 72-h horizon for the evaluated hydrometeor thresholds. This indicates that the proposed physical-space regression and multi-scale spectral supervision help prevent the model from regressing toward the mean, thereby addressing a key bottleneck of standard data-driven models in extreme hydrometeor prediction.

Supplementary Fig. S1 provides a more detailed view of the extreme-value behavior. The radar charts summarize CSI differences relative to GFS across hydrometeor species and percentile thresholds, showing that the improvement is not confined to a single hydrometeor category or threshold choice. The fractions skill score (FSS) \cite{roberts2008scale} heatmaps further evaluate neighborhood-scale spatial agreement at the 97th percentile for representative pressure levels. Entries marked as not applicable (N/A) correspond to species--level combinations whose ERA5 reference values are too sparse for a stable percentile-threshold evaluation. These results indicate that the gains in extreme hydrometeor detection are accompanied by improved spatial coherence at small neighborhood scales, rather than being solely a pointwise threshold effect.

While the operational GFS baseline also remains highly competitive for smooth thermodynamic variables, benefiting from its explicit physical constraints for large-scale balanced flow and radiation, PredHydro-Net demonstrates distinct advantages in predicting highly skewed hydrometeor variables. This highlights the model's complementary strength in resolving complex, zero-inflated atmospheric processes.

Supplementary Fig. S2 expands this comparison for continuous thermodynamic variables by showing global error-difference maps and zonal-mean RMSE profiles for 1000 hPa temperature and specific humidity. The spatial maps highlight regions where PredHydro-Net has lower or higher RMSE than GFS, while the zonal profiles summarize the latitudinal dependence of both models. Some polar regions also exhibit lower errors for PredHydro-Net, though these areas carry relatively small meridional weights in the globally averaged evaluation. Rather, the supplementary analysis helps delineate the scope of the proposed model: its primary advantage lies in highly skewed hydrometeor prediction, whereas GFS remains a strong reference for large-scale circulation and smooth thermodynamic fields, especially in upper-level variables.

\subsection{Global Consistency and Three-Dimensional Structure}

\begin{figure}[!htbp]
\centering
\includegraphics[width=\linewidth]{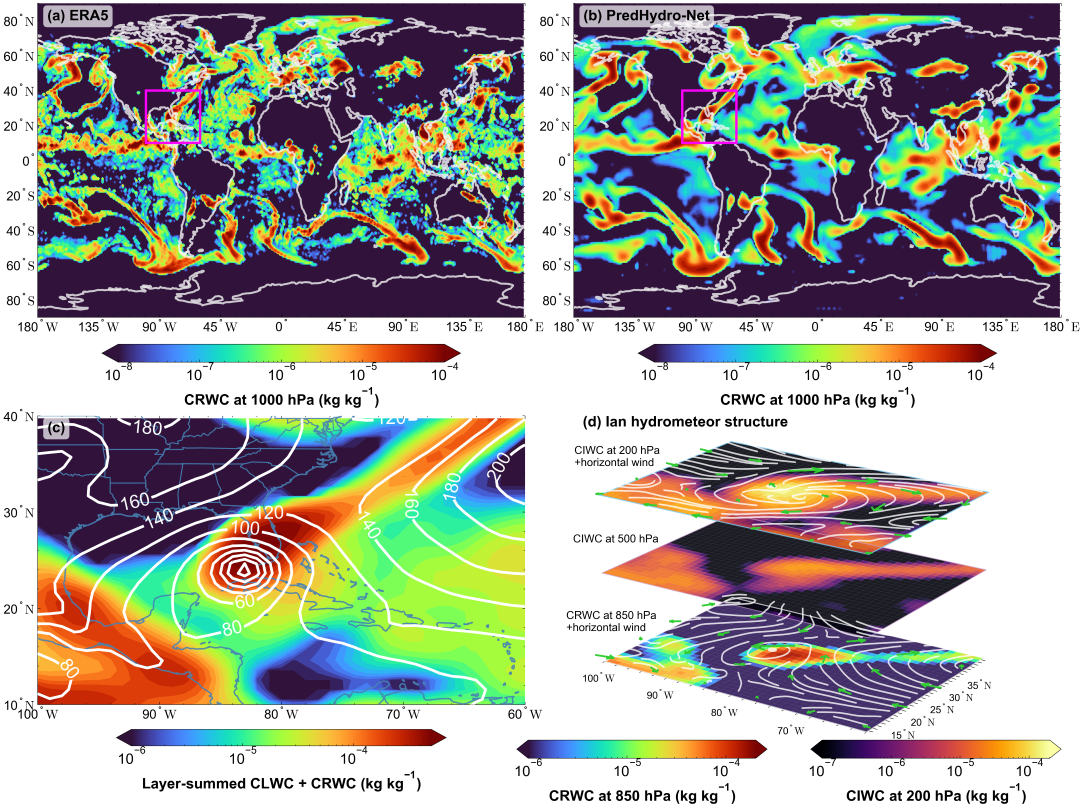}
\caption{Global spatial consistency and three-dimensional hydrometeor structure in an extreme weather case. The Hurricane Ian case is initialized at 00 UTC on 26 September 2022 and evaluated at 00 UTC on 28 September 2022. \textbf{a, b}, Global distributions of specific rain water content (CRWC) at 1000 hPa from the ERA5 reference and the PredHydro-Net prediction, illustrating large-scale spatial organization. \textbf{c}, Synoptic-scale view of Hurricane Ian showing the PredHydro-Net layer-summed liquid hydrometeor content (CLWC + CRWC over the evaluated pressure levels) with overlaid ERA5 1000 hPa geopotential-height contours. \textbf{d}, Three-dimensional close-up of Hurricane Ian, showing specific cloud ice water content (CIWC) at 200 and 500 hPa and CRWC at 850 hPa together with ERA5 horizontal wind vectors. The layered structure illustrates the correspondence between predicted hydrometeors and the surrounding dynamical fields.}
\label{fig:ian_structure}
\end{figure}

For global data-driven models, maintaining large-scale coherence while generating local high-frequency texture is an important challenge. As shown in Fig.~\ref{fig:ian_structure}a, b, the predicted global CRWC field at 1000 hPa crosses the international date line smoothly, aided by circular topological padding, and reasonably reproduces the large-scale morphology of the intertropical convergence zone (ITCZ) and midlatitude storm tracks. This suggests that the generated textures are at least partly constrained by large-scale forcing.

To further examine vertical structural consistency, we analyze Hurricane Ian at the +48 h valid time using regional and three-dimensional slices (Fig.~\ref{fig:ian_structure}c, d). The regional layer-summed CLWC+CRWC field highlights the low- and mid-level liquid hydrometeor envelope around the cyclone, while the ERA5 1000 hPa geopotential-height contours locate the synoptic-scale circulation center. At 200 hPa, the model represents a cloud-ice outflow band associated with upper-level divergence. At 500 hPa, it captures a mixed-phase spiral cloud band. At 850 hPa, regions of intense rain water content broadly correspond to ERA5 cyclonic horizontal winds. This three-dimensional hydrometeor structure, spanning multiple pressure levels and remaining consistent with continuous dynamical fields, supports the design motivation of TQ2HydroFiLM as a mechanism for introducing macroscopic thermodynamic constraints.

\subsection{Spectral Characteristics and High-Frequency Textures}

\begin{figure}[!htbp]
\centering
\includegraphics[width=\linewidth]{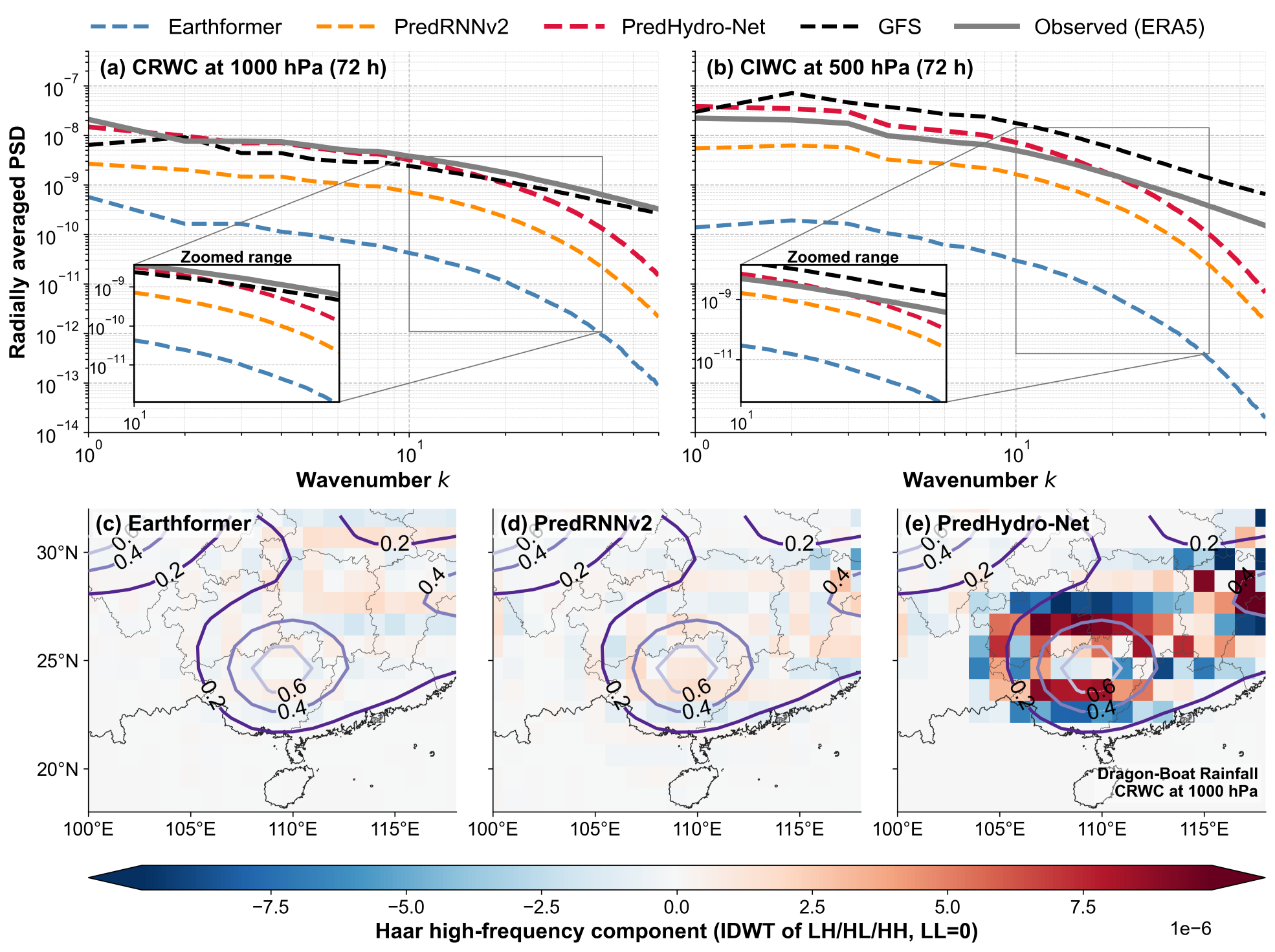}
\caption{Spectral characteristics and high-frequency texture of hydrometeor predictions. \textbf{a, b}, Radially averaged two-dimensional power spectral density (PSD) for CRWC at 1000 hPa and CIWC at 500 hPa at a 72-h lead time, compared with the ERA5 reference and model baselines. \textbf{c--e}, Regional case study of a Dragon-Boat rainfall event in South China, initialized at 00 UTC on 18 June 2022, showing the CRWC 1000 hPa Haar high-frequency component reconstructed by inverse DWT from the low-high/high-low/high-high (LH/HL/HH) subbands with the low-low (LL) subband set to zero. Purple contours denote the ERA5 vertically integrated liquid hydrometeor path (CLWC + CRWC, kg m$^{-2}$), used as a large-scale precipitation-system reference. PredHydro-Net better preserves localized high-frequency structures within the frontal precipitation region than the smoother deep learning baselines.}
\label{fig:spectral_texture}
\end{figure}

Spatial spectra provide an important diagnostic for assessing excessive smoothing. Figure~\ref{fig:spectral_texture}a, b compares the radially averaged power spectral density (PSD) of specific rain water content (CRWC) at 1000 hPa and specific cloud ice water content (CIWC) at 500 hPa. MSE-driven baseline models show high-frequency energy attenuation for wavenumbers $k>10^1$, with spectral curves decreasing too rapidly. By combining Haar DWT decomposition with spectral amplitude matching, PredHydro-Net produces spectra closer to the ERA5 reference over the relevant wavenumber range.

This spectral improvement is also visible in an extreme weather case. Figure~\ref{fig:spectral_texture}c--e shows a Dragon-Boat rainfall event over South China. By extracting the high-frequency texture component of the predicted CRWC 1000 hPa field through inverse Haar reconstruction with $\mathrm{LL}=0$, we observe that baseline models mainly produce broad and smooth precipitation patterns. PredHydro-Net better preserves broken, discrete, and intense convective-cell structures within the frontal region indicated by the ERA5 liquid hydrometeor path contours, suggesting that frequency-domain constraints help represent high-frequency spatial structure.

\subsection{Global Climatological Consistency}

\begin{figure}[!htbp]
\centering
\includegraphics[width=\linewidth]{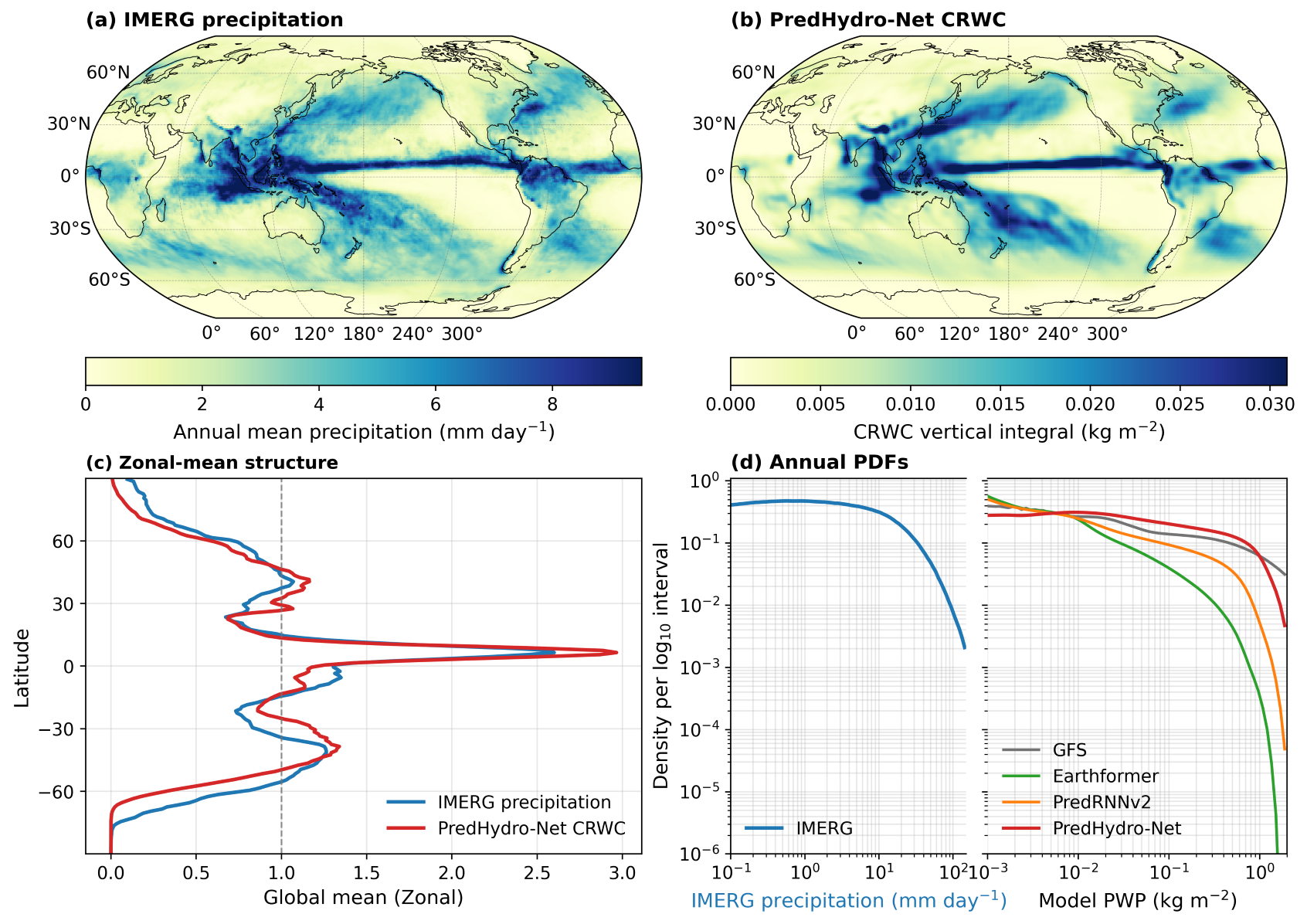}
\caption{Annual-mean climatological comparison between Integrated Multi-satellitE Retrievals for GPM (IMERG) and model forecasts. \textbf{a}, IMERG annual mean precipitation rate (mm day$^{-1}$) in Robinson projection. \textbf{b}, PredHydro-Net annual mean rain water path (RWP, vertically integrated CRWC, $\text{kg m}^{-2}$). \textbf{c}, Zonal-mean structure normalized by the global mean, comparing IMERG precipitation and PredHydro-Net CRWC vertical integral. \textbf{d}, Annual probability density functions shown on separate x-axes: IMERG precipitation (left, $\text{mm day}^{-1}$) and model total precipitating water path (PWP = CRWC + CSWC, right, $\text{kg m}^{-2}$) for GFS, Earthformer, PredRNNv2, and PredHydro-Net.}
\label{fig:imerg_reference}
\end{figure}

Beyond pointwise error metrics, we assess the large-scale climatological structure of PredHydro-Net using the the Global Precipitation Measurement (GPM) mission's Integrated Multi-satellitE Retrievals for GPM (IMERG) product satellite precipitation estimates \cite{hou2014global,huffman2020integrated} as an independent reference. Because ERA5-based training may not explicitly constrain global precipitation patterns, a comparison with IMERG provides a complementary view of the model's climatological behavior.

Figure~\ref{fig:imerg_reference}a, b shows the annual-mean Robinson projections of IMERG precipitation and the PredHydro-Net CRWC vertical integral. Although satellite precipitation and model hydrometeor content are not identical physical quantities, the comparison provides an independent check of whether the modeled rain-water activity occurs in climatologically plausible regions. The predicted CRWC field reproduces the major climatological features, including the intertropical convergence zone (ITCZ), monsoon systems, and midlatitude storm-track regions in both hemispheres. The zonal-mean profile (Fig.~\ref{fig:imerg_reference}c) confirms that the normalized meridional structure of the model's CRWC vertical integral broadly follows the IMERG precipitation signature, particularly in tropical and subtropical latitudes.

To ensure a rigorous evaluation against IMERG, we apply a tailored comparison strategy. For global spatial morphology (Fig.~\ref{fig:imerg_reference}b, c), we compare IMERG with the model's liquid rain water path (vertically integrated CRWC) to avoid the high uncertainties associated with satellite solid-precipitation retrievals at higher latitudes. Conversely, for the global statistical distribution (Fig.~\ref{fig:imerg_reference}d), we evaluate the total precipitating water path (PWP, integrating both CRWC and CSWC). This ensures that the model's probability density function captures the complete spectrum of global precipitation events, matching the comprehensive aggregation of the IMERG product. As shown in Fig.~\ref{fig:imerg_reference}d, PredHydro-Net produces a broader and more heavy-tail-rich PWP distribution than Earthformer and PredRNNv2 under the same ERA5 training setup, indicating that physical-space regression and spectral supervision help maintain more realistic extreme-value behavior. The boreal winter (December-January-February, DJF) / boreal summer (June-July-August, JJA) comparison in Supplementary Fig. S3 further confirms that the model captures the seasonal migration of major precipitation systems.

\subsection{Disentangling Physical Forcing from Stochastic Textures}

In deep learning-based weather prediction, generative adversarial networks (GANs) and diffusion models are often questioned because their high-frequency textures may contain stochastic hallucinations. To provide an initial interpretation of PredHydro-Net, we combine regional physical-consistency diagnostics with gradient-based attribution, formulated as the signed Input$\times$Gradient score \cite{shrikumar2017learning} $\mathbf{S} = (\mathbf{x} - \mathbf{x}_0) \odot \nabla_{\mathbf{x}} y$, to analyze whether extreme hydrometeor generation is connected to physically meaningful input structures.

\begin{figure}[!htbp]
\centering
\includegraphics[width=\linewidth]{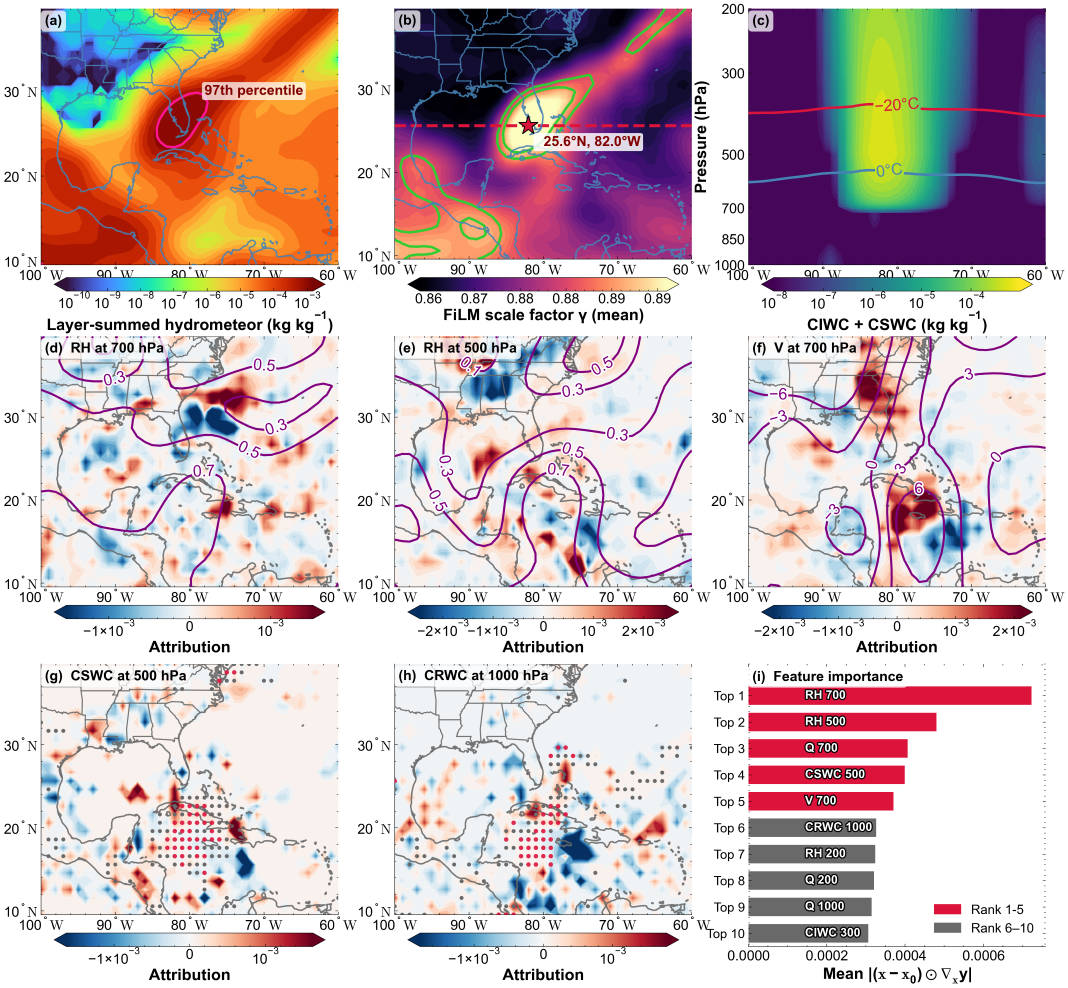}
\caption{Physical interpretability and feature attribution based on gradient analysis for Hurricane Ian, initialized at 00 UTC on 26 September 2022. \textbf{a}, Spatial distribution of the target hydrometeor field, with extremes defined by the 97th percentile within the Hurricane Ian region. The target field is the layer-summed hydrometeor specific content over the evaluated species and pressure levels. \textbf{b}, Channel-mean FiLM scale factor $\gamma$ over the same region, with the dashed line indicating the latitude used for the cross-section. \textbf{c}, Vertical cross-section of ice-phase hydrometeors (CIWC + CSWC) with 0$^{\circ}$C and $-20^{\circ}$C isotherms. \textbf{d--h}, Signed spatial attribution maps for selected input channels, computed from $(\mathbf{x} - \mathbf{x}_0) \odot \nabla_{\mathbf{x}} y$ with a temporal-mean baseline $\mathbf{x}_0$. Panels \textbf{d--f} use SmoothGrad for continuous variables, whereas panels \textbf{g--h} use raw Input$\times$Gradient for sparse hydrometeor inputs. Purple contours show the corresponding ERA5 continuous input fields, and gray/red dots mark high-percentile sparse hydrometeor regions. \textbf{i}, Feature-importance ranking based on the mean absolute attribution within the target mask.}
\label{fig:attribution}
\end{figure}

Figure~\ref{fig:attribution}a--c first examines whether the selected high-risk hydrometeor region is physically organized before applying attribution analysis. The 97th-percentile (P97) target mask in Fig.~\ref{fig:attribution}a isolates the intense hydrometeor core around Hurricane Ian rather than a randomly selected texture region. The corresponding FiLM response in Fig.~\ref{fig:attribution}b is spatially concentrated near the same storm-scale envelope, indicating that thermodynamic-branch modulation is strongest where hydrometeor generation is active. The vertical cross-section in Fig.~\ref{fig:attribution}c provides a further physical-consistency check: ice-phase hydrometeors are organized above the lower-tropospheric liquid region and are aligned with the 0$^{\circ}$C and $-20^{\circ}$C isotherms, consistent with the expected mixed-phase and ice-cloud structure of a tropical cyclone. These panels therefore establish that the analyzed target corresponds to a meteorologically coherent system, rather than an isolated high-frequency artifact.

To provide model interpretability,\cite{kashinath2021physics,chen2025interpretable,bracco2025machine,wu2023interpretable,mamalakis2026unraveling} we compute spatial attribution maps for the P97 hydrometeor target within the Hurricane Ian region, as shown in Fig.~\ref{fig:attribution}d--h. The attribution is based on $\mathbf{S} = (\mathbf{x} - \mathbf{x}_0) \odot \nabla_{\mathbf{x}} y$, where $\mathbf{x}_0$ is a temporal-mean baseline, $y$ is the scalar hydrometeor target averaged over the P97 mask, and $\odot$ denotes the element-wise Hadamard product. For continuous inputs, we additionally use SmoothGrad \cite{smilkov2017smoothgrad} to reduce pixel-scale gradient noise, and channel ranking is based on the mean absolute attribution $|\mathbf{S}|$ within the target mask. The high-attribution responses \cite{gunning2019xai,minh2022explainable} are not purely random scatter points. Instead, they concentrate near large-scale structures associated with heavy precipitation systems, such as cyclones and fronts. This suggests that PredHydro-Net's extreme hydrometeor prediction is linked to the macroscopic thermodynamic and dynamic background.

Under the TQ2HydroFiLM cross-guidance architecture, the model tends to use continuous thermodynamic and dynamic fields as a macroscopic envelope. PatchGAN \cite{isola2017image} mainly supplements high-frequency texture within these high-risk regions. Therefore, the attribution results are consistent with the physical understanding that hydrometeor occurrence and extremes are constrained by large-scale meteorological conditions.

\subsection{Thermodynamic and Kinematic Drivers of Hydrometeor Extremes}

To further analyze the model's dependence on input features, we rank all input channels by the mean absolute attribution within the target mask (Fig.~\ref{fig:attribution}i). The top-ranked features are broadly consistent with classical precipitation-formation theory.

\textbf{Relative humidity at 700 and 500 hPa.} Among the 93 input channels, mid- and lower-tropospheric relative humidity (RH 700 and RH 500) ranks highly in contribution to hydrometeor generation. Physically, RH implicitly combines moisture availability with the temperature constraint described by the Clausius-Clapeyron relationship.\cite{romps2014analytical,cox2015humidity,lenderink2017super} Near-saturation in the middle troposphere is an important condition for deep moist convection and cloud droplet growth. This result suggests that explicitly introducing RH, rather than relying only on specific humidity $q$, provides a direct physical prior for phase transitions and makes the model prediction consistent with the thermodynamic understanding that unsaturated environments are less favorable for condensation.

\textbf{Dynamic lifting and moisture transport.} The relatively high attribution weights of 700-hPa meridional wind ($v$) and 1000-hPa specific humidity ($q$) indicate that the model's extreme hydrometeor prediction depends on dynamic transport and low-level moisture supply. Sufficient low-level moisture provides energy and material sources for convective systems, while low-level jets or frontal convergence can provide forced ascent. The appearance of wind-related gradients suggests that PredHydro-Net may capture signals associated with the synergistic forcing of moisture advection and dynamic lifting.

\textbf{Autoregressive hydrometeor memory.} Historical hydrometeor fields, such as CSWC at 500 hPa and CRWC at 1000 hPa, also rank highly. Physically, this is related to the advective memory of cloud systems and to microphysical sedimentation processes, such as the conversion and fallout of cloud ice and snow into rain.

Overall, PredHydro-Net performs well in statistical metrics, and its attribution patterns in this representative case study are consistent with the classical convective-evolution chain from low-level moisture supply to dynamic convergence and lifting, followed by mid-level saturation and condensation.

\subsection{Component Ablation and Trade-off Analysis}

\begin{figure}[!htbp]
\centering
\includegraphics[width=\linewidth]{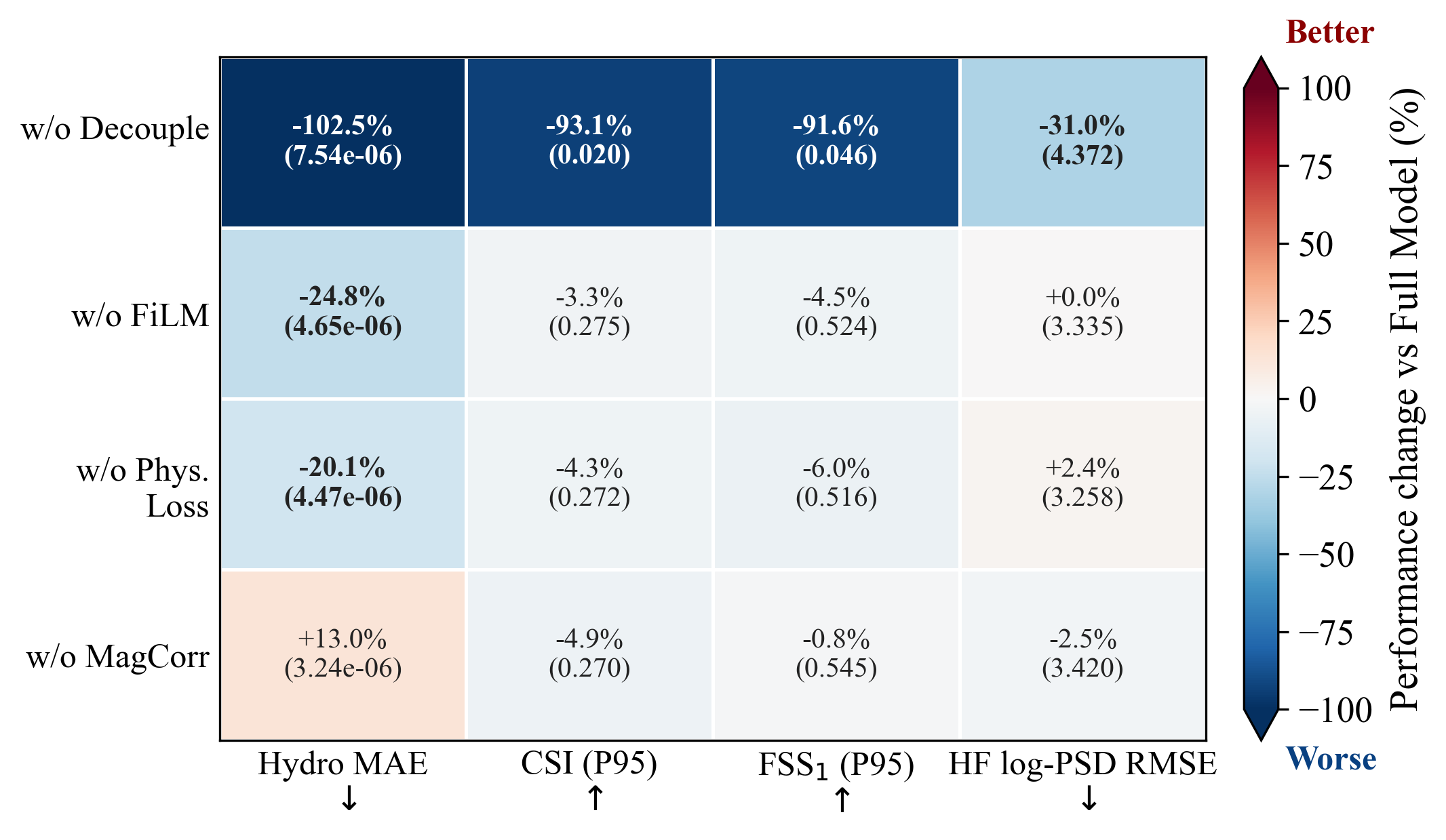}
\caption{Component ablation study. Performance change relative to the full PredHydro-Net model when individually removing key design components: \textit{w/o Decouple} (single-decoder baseline replacing the dual-decoder structure), \textit{w/o FiLM} (no TQ2HydroFiLM cross-branch modulation), \textit{w/o Phys.~Loss} (no physical-space regression penalty), and \textit{w/o MagCorr} (no magnitude-correction term). Each cell reports the signed performance change relative to the full model, with the absolute metric value shown in parentheses. Metrics include hydrometeor mean absolute error (MAE), CSI at the 95th percentile (P95), neighborhood-scale FSS$_1$(P95), and high-frequency log-PSD RMSE. Positive values indicate improvement relative to the full model after accounting for the metric direction. Negative values indicate degradation.}
\label{fig:ablation}
\end{figure}

To isolate the contribution of individual design components, we conduct ablation experiments in which each key element is individually removed from the full PredHydro-Net configuration. Figure~\ref{fig:ablation} summarizes the relative performance change against the full model across four metrics: hydrometeor mean absolute error (MAE), CSI at the 95th percentile (P95), neighborhood-scale FSS$_1$(P95), and high-frequency log-PSD RMSE. The primary purpose of this analysis is to test the coupled architectural hypothesis behind PredHydro-Net: hydrometeor prediction benefits first from separating dense thermodynamic reconstruction from sparse hydrometeor generation, and then from using the thermodynamic branch to provide one-way physical conditioning through TQ2HydroFiLM.

The most significant degradation occurs when the dual-decoder decoupling structure is removed (\textit{w/o Decouple}): hydrometeor MAE increases by approximately 102.5\% and CSI (P95) and FSS$_1$ (P95) decrease by 93.1\% and 91.6\%, respectively. This collapse indicates that the decoupled thermodynamic and hydrometeor decoders are not only an implementation detail, but the enabling structure that prevents dense continuous-variable learning from overwhelming the sparse hydrometeor objective. Removing the TQ2HydroFiLM cross-branch modulation while retaining the decoders (\textit{w/o FiLM}) still leads to a 24.8\% increase in MAE and reductions in CSI and FSS. The two ablations therefore show a hierarchical effect: decoder decoupling establishes a separate hydrometeor pathway, whereas FiLM makes that pathway physically conditioned by the predicted thermodynamic envelope. The much larger failure of the non-decoupled baseline further suggests that FiLM is effective only after the branches have been separated, rather than as a standalone post-hoc correction.

The loss-related ablations show a complementary pattern. Removing the physical-space regression loss (\textit{w/o Phys.~Loss}) causes a 20.1\% degradation in MAE and noticeable reductions in both CSI and FSS, supporting the need to supervise hydrometeors in their original physical units. Removing the magnitude-correction term (\textit{w/o MagCorr}) lowers hydrometeor MAE but weakens CSI, FSS, and high-frequency spectral accuracy. We therefore interpret this term as a trade-off component rather than a uniformly beneficial accuracy term: it helps preserve extreme-event and high-frequency structure at the cost of a small increase in global mean error. Together, these results indicate that the core gain comes from the cascaded Decouple--FiLM design, while the loss terms regulate the balance between mean error minimization and physically meaningful hydrometeor structure.

\subsection{Parameter Sensitivity and Pareto Selection}

\begin{figure}[!htbp]
\centering
\includegraphics[width=\linewidth]{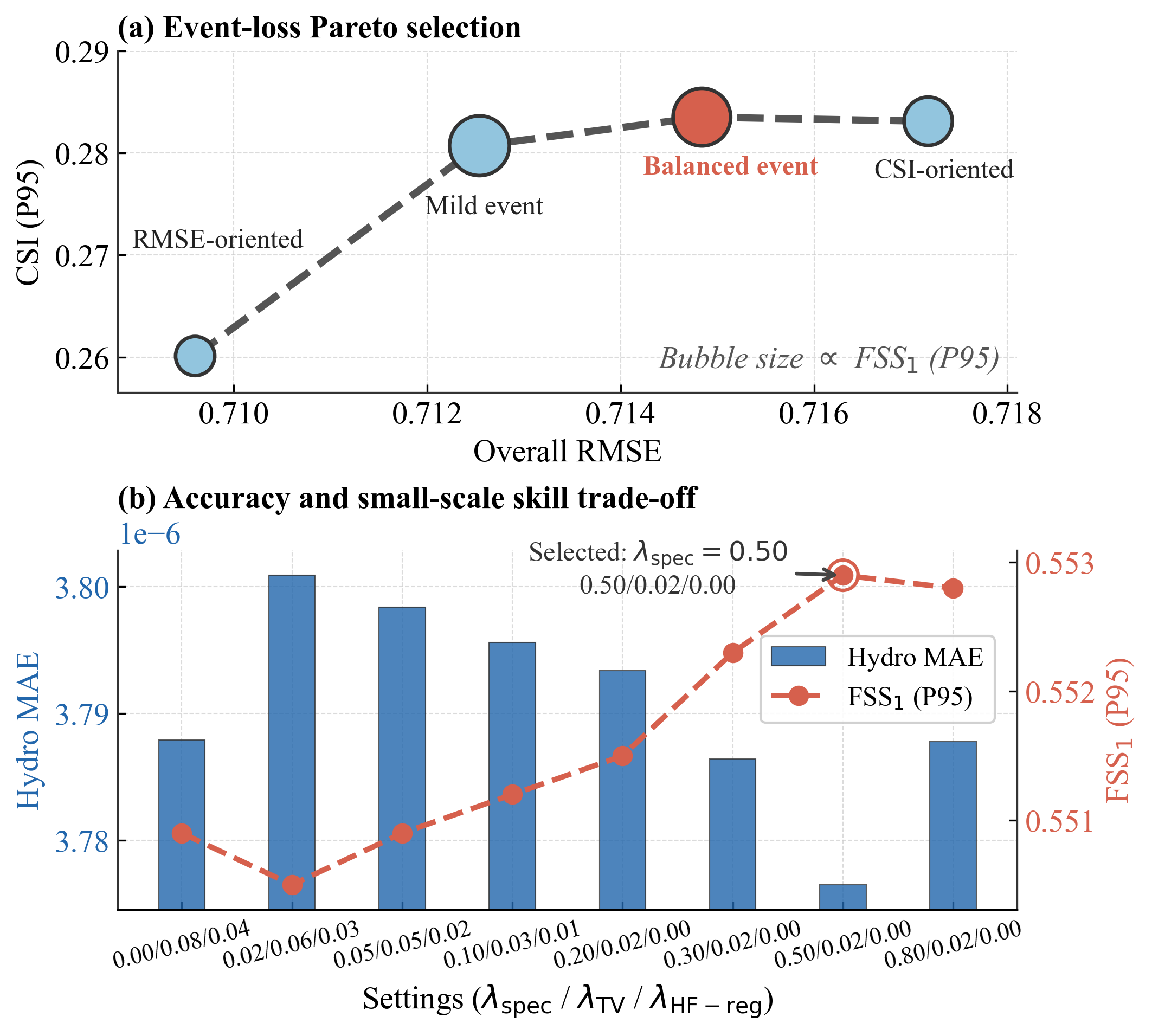}
\caption{Parameter sensitivity and Pareto selection for physical and spectral supervision. \textbf{a}, Event-loss Pareto selection: trade-off between overall RMSE and CSI at the 95th percentile (P95), with bubble size proportional to FSS$_1$(P95). The point labels denote increasing event penalty strength from RMSE-oriented to CSI-oriented settings, with the Balanced event setting selected as the compromise between average accuracy and extreme-event detection. \textbf{b}, Sensitivity of the combined spectral, total-variation (TV), and high-frequency regularization (HF-reg) weights ($\lambda_{\rm spec}/\lambda_{\rm TV}/\lambda_{\rm HF\text{-}reg}$) on hydrometeor MAE (bars, left axis) and nearest-neighbor FSS$_1$(P95) (line, right axis). The selected setting (0.50/0.02/0.00) maximizes FSS$_1$(P95) while maintaining a low hydrometeor MAE.}
\label{fig:pareto_spectral}
\end{figure}

In generative weather prediction, optimizing for lower RMSE may suppress the model's ability to represent extreme events. To identify a suitable operating point, we conduct parameter-sensitivity experiments with two groups of hyperparameters. For the event-loss sweep in Fig.~\ref{fig:pareto_spectral}a, the four points differ only in their loss weighting strategies, transitioning from an RMSE-oriented baseline to a strict CSI-oriented configuration. Specifically, we progressively increased the penalties for underprediction and high-intensity events, along with stronger adversarial and physical hydrometeor weights, while keeping the core architecture fixed. Exact hyperparameter configurations for each setting are detailed in the Methods section.

\textbf{Physical-penalty trade-off.} We examine how asymmetric physical-loss weights affect model performance. As shown in Fig.~\ref{fig:pareto_spectral}a, increasing the penalty strength from the RMSE-oriented setting to the Balanced-event setting raises the extreme-precipitation CSI (P95) from 0.260 to 0.284 while maintaining a relatively high FSS$_1$ (P95). This improvement is accompanied by a modest increase in overall RMSE from 0.710 to 0.715, indicating a trade-off between average accuracy and extreme event detection. Moving further to the CSI-oriented setting provides little additional CSI (P95) gain and reduces neighborhood-scale FSS, suggesting that overly strong event penalties can sacrifice spatial continuity to improve pointwise extreme-value hits. Based on this analysis, we adopt the Balanced event setting as the physical-constraint compromise baseline.

\textbf{Spectral regularization sensitivity.} We then evaluate the combined sensitivity of the spectral amplitude loss and high-frequency regularization weights ($\lambda_{\rm spec}/\lambda_{\rm TV}/\lambda_{\rm HF\text{-}reg}$). TV denotes total-variation regularization and HF-reg denotes high-frequency regularization. As shown in Fig.~\ref{fig:pareto_spectral}b, increasing $\lambda_{\rm spec}$ generally improves nearest-neighborhood skill FSS$_1$ (P95), with the response saturating at high spectral weights. The selected setting of 0.50/0.02/0.00 reaches the highest FSS$_1$ (P95) and keeps hydrometeor MAE among the lowest tested values, confirming that spectral constraints can complement standard regression losses in mitigating spatial smoothing without degrading average hydrometeor accuracy.

\section{Discussion}

Accurate global three-dimensional hydrometeor prediction is an important challenge for numerical weather prediction and climate-system modeling. Because hydrometeor variables are strongly zero-inflated and long-tailed, MSE-optimized deep learning models tend to regress toward the mean and smooth the physical structure and high-frequency texture of precipitation systems. PredHydro-Net addresses this problem by combining a decoupled thermodynamic--hydrometeor architecture, TQ2HydroFiLM physical modulation, wavelet-based multi-scale supervision, spectral amplitude matching, adversarial texture generation, and physical-space regression. Together, these components separate macroscopic physical envelopes from high-frequency texture components while maintaining a link to physically meaningful thermodynamic and dynamic drivers.

The sensitivity and ablation analyses show that PredHydro-Net improves extreme hydrometeor and high-frequency texture representation while maintaining relatively stable large-scale forecast errors. In the ERA5-referenced 72-h evaluation, PredHydro-Net improves selected hydrometeor RMSE and extreme-threshold CSI metrics relative to the deep learning baselines and the uniformly processed GFS forecast. It also reproduces radial power spectra and coordinated three-dimensional hydrometeor structures in selected typhoon and monsoon rainfall cases. Gradient-based attribution further shows that the model is sensitive to relative humidity and dynamic convergence, consistent with basic physical understanding of hydrometeor formation.

Several avenues for future refinement remain. First, because global three-dimensional grids with 93 physical channels impose substantial memory and computational costs, the model is trained at $1^{\circ}$ spatial resolution using five years of ERA5 data. As suggested by standardized benchmarks such as WeatherBench, this intermediate-resolution setting provides a practical balance between computational feasibility and dynamical evaluation for testing global modeling frameworks.\cite{rasp2020weatherbench,rasp2024weatherbench} Future work will extend the data scale and introduce parallel partitioning to move toward higher resolutions, such as $0.25^{\circ}$.

Second, the model mainly focuses on intense hydrometeor cores, such as values above the 95th percentile, and their large-scale topology. In this multi-objective optimization setting, some weak, fragmented hydrometeor details may be lost. Weather prediction is not identical to image super-resolution. In atmospheric physics, latent heating in intense convective regions plays an important role in circulation feedback. Therefore, prioritizing dynamically meaningful extreme structures under smoothing constraints is an interpretable compromise.

A related limitation is that extreme hydrometeor targets inherit uncertainty from the reference data. Unlike smoother thermodynamic variables, cloud and precipitation-related hydrometeor fields are highly intermittent and strongly skewed, so their most intense values in reanalysis products or satellite retrievals may be affected by larger representation, assimilation, or retrieval uncertainties. Such uncertainty may impose an upper bound on deterministic skill, especially when regression and adversarial objectives are trained against a single reference realization. Future work could incorporate uncertainty quantification or probabilistic generative models to better distinguish predictable physical signals from reference-data uncertainty in extreme hydrometeor prediction.

Third, the base spatiotemporal evolution framework is autoregressive and based on PredRNNv2. For the 72-h lead time considered here, this architecture combined with generative-adversarial inference is computationally efficient and yields controlled errors. For longer lead times, however, recurrent architectures may suffer from error accumulation. Future work will explore Transformer- or graph neural network-based global architectures to better represent large-scale thermodynamic boundary conditions, and will consider denoising diffusion probabilistic models as alternatives to adversarial generation for more stable conditional generation.

Finally, comparisons between data-driven models and operational NWP systems naturally entail intrinsic differences in underlying cloud microphysics schemes, variable definitions, and vertical discretization. While evaluating against the operational European Centre for Medium-Range Weather Forecasts (ECMWF) Integrated Forecasting System (IFS) would provide the most direct baseline for an ERA5-trained model, full three-dimensional, operational hydrometeor forecasts are not as conventionally archived or openly accessible as those from the NCEP GFS. Future comprehensive evaluations will benefit from expanding the baseline pool to include consistently processed hindcasts from multiple operational centers as these specialized hydrometeor datasets become more publicly available.

Overall, PredHydro-Net provides a physics-informed deep learning framework for mitigating smoothing in long-tailed atmospheric-variable prediction. The results suggest that embedding physical constraints and frequency-domain supervision into neural architectures is a feasible direction for future high-resolution weather models that more explicitly represent cloud microphysical processes.

\section{Methods}

\subsection{Dataset and Feature Engineering}

This study uses the ERA5 reanalysis from the European Centre for Medium-Range Weather Forecasts (ECMWF) as the reference dataset \cite{hersbach2020era5} for model training and evaluation. The dataset spans five years and is obtained as 6-hourly NetCDF files on a regular $1^{\circ} \times 1^{\circ}$ latitude-longitude grid. No additional horizontal interpolation is applied in this study. The raw latitude grid includes both polar endpoints. To facilitate tensor-based global convolution, the southern polar endpoint row is excluded, yielding a uniform $180\times360$ latitude-longitude spatial dimension for both inputs and targets. To represent the three-dimensional atmospheric environment, we construct 93 input channels, consisting of 92 ERA5-derived meteorological channels and one appended latitude channel (Supplementary Table S1).

The temporal split is chronological: samples initialized during 2018--2021 are used for model training, and samples initialized in 2022 are held out for independent evaluation. Each sample uses two past 6-hourly atmospheric states as input and predicts twelve future 6-hourly states, corresponding to lead times from +6 h to +72 h. Accordingly, the model input tensor has shape $\mathbf{X}\in\mathbb{R}^{B\times2\times93\times180\times360}$, and the supervised output tensor has shape $\mathbf{Y}\in\mathbb{R}^{B\times12\times30\times180\times360}$, where the 30 target channels consist of thermodynamic variables and hydrometeor variables on the evaluated pressure levels. The normalization statistics and input preprocessing transforms are estimated from the training years and then applied unchanged to the 2022 evaluation set.

When spatial averages are computed in training objectives and global diagnostics on the regular latitude--longitude grid, grid cells are weighted by the cosine of latitude to approximate area-weighted global means and avoid overemphasizing polar grid rows, following common practice in global data-driven weather modeling.\cite{chen2023fuxi}

\textbf{Basic state variables} (35 channels) include geopotential ($z$), temperature ($T$), specific humidity ($q$), and horizontal wind components ($u, v$) on seven pressure levels: 200, 300, 500, 700, 850, 925, and 1000 hPa.

\textbf{Hydrometeor state variables} (28 channels) include four ERA5 specific content variables: specific cloud ice water content (CIWC), specific cloud liquid water content (CLWC), specific rain water content (CRWC), and specific snow water content (CSWC), extracted on the same seven pressure levels.

\textbf{Physical diagnostic features} (21 channels) include relative vorticity, divergence, and RH at all seven levels. RH is explicitly computed and included as an important physical quantity related to cloud formation and phase transition.\cite{haag2003freezing}

\textbf{Structural and geometric features} (9 channels) include eight layer-thickness variables between selected pressure levels, such as 925--1000 hPa and 500--1000 hPa, to represent atmospheric stability and vertical temperature gradients. A latitude channel is appended separately to account for global geometric effects.

The seven-level input design provides additional vertical context, including 300 and 925 hPa fields that help describe upper-tropospheric and boundary-layer structure. The prediction targets are defined on five standard pressure levels (200, 500, 700, 850, and 1000 hPa) to focus the supervised task on levels with stable evaluation statistics and clear physical interpretation, while reducing the computational cost of global three-dimensional prediction.

To improve training stability across variables with different distributions, we apply an input-side quantile transform \cite{wood2004hydrologic,pedregosa2011scikit} to map predictor variables toward a normal distribution. Because hydrometeor input variables are strongly zero-inflated and skewed, a natural-log transformation, $y=\ln(1+x)$, is applied before the quantile transform to reduce the dominance of zero values and stabilize variance across orders of magnitude. This preprocessing is used for the 93 input channels and is distinct from the output normalization used for supervised losses.

An independent satellite-based precipitation reference is additionally derived from the IMERG product.\cite{hou2014global,huffman2020integrated} The native IMERG product is provided at $0.1^{\circ}$ spatial resolution and half-hourly temporal resolution. In this study, it is aggregated to daily precipitation rates on the same $1^{\circ}\times1^{\circ}$ grid for climatological comparison. IMERG is used only for climatological consistency analyses of precipitation-related spatial patterns and seasonal migration, and is not used for model training, normalization, or hyperparameter selection.

\subsection{Baselines}

To evaluate PredHydro-Net, we compare it with both an NWP baseline and representative deep learning baselines.

\textbf{Operational GFS.} The NCEP GFS is used as an operational NWP reference.\cite{han2017updates} The GFS files are also obtained at 6-hourly forecast lead times on a regular $1^{\circ} \times 1^{\circ}$ latitude-longitude grid, and no additional horizontal interpolation is applied before comparison. The fields are evaluated on the same $180\times360$ grid as the ERA5-based targets and model outputs. GFS and ERA5 hydrometeor variables are not stored in exactly the same thermodynamic basis, so we first convert the exported GFS hydrometeor dry-air mixing ratios to ERA5-style moist-air specific contents before computing metrics. For each pressure level, the conversion follows the thermodynamic relationship:
\begin{equation}
q_{k}^{\rm GFS} = \frac{r_{k}^{\rm GFS}}{1 + r_{v}^{\rm GFS} + \sum_{i} r_{i}^{\rm GFS}} ,
\end{equation}

where $r_{k}^{GFS}$ is the GFS dry-air mixing ratio of a specific hydrometeor species $k$, $r_{v}^{GFS}$ is the GFS water vapor mixing ratio, and the summation in the denominator accounts for all modeled hydrometeor species at that level. After conversion, hydrometeor values below $10^{-8}\text{ kg kg}^{-1}$ are cleanly truncated to zero to eliminate numerical background noise.

\textbf{Earthformer and PredRNNv2.} These models are used as representative spatiotemporal sequence-prediction baselines. To ensure a fair and rigorous comparison, both baseline models are configured with optimized hyperparameters tailored to this global high-resolution objective, aiming to maintain competitive forecast skill while mitigating their inherent over-smoothing tendencies.

For species-specific profile and threshold diagnostics, we report only the pressure levels with physically meaningful hydrometeor occurrence. Several upper-level liquid or rain-water channels, especially CRWC above the lower troposphere, are almost always zero or many orders of magnitude smaller than lower-level values, so they provide unstable percentile thresholds and little meaningful forecast signal.

\subsection{Model Architecture}

To account for global topology and reduce convolutional boundary artifacts, we introduce global padding before spatial encoding. Standard planar convolution does not recognize the periodicity of longitude and can therefore introduce unphysical discontinuities near the 0$^{\circ}$/360$^{\circ}$ seam. We therefore apply circular padding of $\pm 48$ grid points in the longitude direction. This margin is chosen to exceed the approximate effective receptive field of the two-stage encoder, four-layer ST-LSTM stack, and decoder (about 35 grid cells), so that convolution-induced edge contamination remains within the auxiliary padded region and is removed by cropping. In the latitude direction, replicate padding is applied with 4 grid points at the northern boundary and 8 grid points at the southern boundary to obtain a size compatible with the downsampling--upsampling hierarchy. The global grid is expanded from $180 \times 360$ to $192 \times 456$, encoded to an internal $48 \times 114$ latent grid, decoded back to $192 \times 456$, and cropped back to $180 \times 360$ before the final losses are computed. Together with an explicit seam-consistency loss over all output channels, this procedure reduces boundary and seam artifacts in the valid physical domain.

To mitigate possible gradient conflict and negative transfer between dense thermodynamic variables and zero-inflated hydrometeor variables, PredHydro-Net uses a decoupled dual-decoder architecture after the ST-LSTM extracts a global latent state on the padded $48 \times 114$ grid. The thermodynamic temperature--humidity branch (TQ Decoder) reconstructs smooth large-scale fields such as temperature and specific humidity, using transposed convolutions for upsampling. The hydrometeor branch (Hydro Decoder) generates precipitation-related fields with higher-frequency structure. To reduce checkerboard artifacts that may arise from standard deconvolution, this branch uses bilinear upsampling followed by standard convolution.

The TQ2HydroFiLM module introduces unidirectional physical modulation. To constrain hydrometeor generation by the large-scale thermodynamic envelope, we insert a FiLM module \cite{perez2018film} at the first upsampling stage of the two decoders, reflecting the strong thermodynamic control on deep convection and cloud formation.\cite{davis2013thermodynamic,rasmussen2020changes} The module extracts intermediate thermodynamic features $F_{\rm TQ}$ and generates a scaling factor $\gamma = \sigma(\mathbf{W}_\gamma F_{\rm TQ})$ and a bias factor $\beta = \mathbf{W}_\beta F_{\rm TQ}$, which are applied to hydrometeor features $F_{\rm hydro}$ through channel-wise affine modulation:
\begin{equation}
F'_{\rm hydro} = \gamma \odot F_{\rm hydro} + \beta ,
\end{equation}
where $\odot$ denotes the element-wise Hadamard product, $\mathbf{W}_\gamma, \mathbf{W}_\beta$ are learnable weight matrices, and $\sigma$ is a sigmoid activation used to limit abnormal amplification. In the implementation, a stop-gradient operation is applied to $F_{\rm TQ}$ to ensure unidirectional information flow, which allows temperature and humidity features to constrain cloud and precipitation generation, while reducing the influence of sparse high-frequency hydrometeor losses on the thermodynamic decoder. It therefore mitigates the multi-objective trade-off in joint optimization.

\subsection{Loss Functions and Optimization}

To mitigate smoothing caused by MSE losses, we design a multi-scale supervision scheme for the hydrometeor branch.

\textbf{Low-frequency envelope constraint with Haar DWT.} A Haar DWT \cite{mallat1989theory} head is used as an auxiliary decoder branch to predict low- and high-frequency hydrometeor subbands. Because extreme precipitation is often underestimated, we apply an asymmetric L1 loss to the predicted low-low (LL) subband and assign a larger penalty ($\times 2.0$) to missed-event regions. This guides the model to better locate large-scale convective systems. Sparse regularization is also applied to the predicted high-frequency subbands (LH, HL, HH) to suppress unrealistic random high-frequency noise.

\textbf{2D FFT spectral amplitude matching.} To improve texture fidelity, we introduce a frequency-domain matching penalty between the prediction and the reference field. A two-dimensional fast Fourier transform (2D FFT) is applied to the spatial fields, and the $L_1$ distance between their normalized amplitude spectra is computed:
\begin{equation}
L_{\rm spectral} = \big\| |\mathcal{F}(Y_{\rm hydro})| - |\mathcal{F}(\hat{Y}_{\rm hydro})| \big\|_1 .
\end{equation}
This loss encourages the generated field to approach the reference energy distribution across wavenumbers, especially in high-frequency ranges, and helps improve the representation of broken cloud structures within fronts and cyclones.

\textbf{PatchGAN adversarial generation.} On top of these physical constraints, a Least Squares generative adversarial network (LSGAN)-style pixel-space discriminator~\cite{mao2017least} is applied only to hydrometeor channels to further supplement high-frequency spatial details.

In atmospheric deep learning, directly optimizing variables with very different magnitudes, from $10^2$ for temperature to $10^{-6}$ for hydrometeor content, can cause gradient-scale imbalance. However, relying too heavily on nonlinear distribution mappings may also compress gradients for non-Gaussian extremes and weaken the representation of intense precipitation cores. To mitigate this trade-off, PredHydro-Net adopts a dual-space optimization framework for the 30-channel outputs that combines differentiable grouped normalization with physical-space penalties.

\textbf{Base regression in standardized space.} During decoding and base loss computation, the model operates in standardized output space for numerical stability. A grouped normalization module (DiffNorm) applies different transformations to different target groups. For continuous thermodynamic variables ($T$, $Q$), Z-score normalization is applied in physical units. For zero-inflated, non-Gaussian hydrometeor variables, a logarithmic transform followed by Z-score standardization is applied:
\begin{equation}
Y_{\rm std} = \frac{\ln(1+Y) - \mu}{\sigma} ,
\end{equation}
where $Y$ represents the physical hydrometeor content, and $\mu, \sigma$ are the channel-wise mean and standard deviation computed over the transformed training set. Most base regression losses, including multi-scale asymmetric L1 losses and the seam-consistency loss at the 0-degree meridian, are computed in this smoother standardized output space to promote stable convergence. This DiffNorm transform is differentiable and is separate from the input-side quantile transform described above.

\textbf{Physical-space regression via differentiable inverse transform.} To improve the representation of extreme convective systems and reduce attenuation caused by latent-space mapping, we introduce a physical-space regression constraint. During backpropagation, standardized decoder outputs are mapped back to their original physical units ($\text{kg kg}^{-1}$) through a differentiable inverse-transform operator. In this physical space, the model computes a modified natural-log Log-L1 loss for hydrometeors and a physical L1 loss for thermodynamic variables. This dual-space constraint, referred to hereafter as Magnitude Correction (MagCorr), benefits from stable standardized-space learning while adding physical-space penalties for missed or false extreme hydrometeor events, leading to a trade-off between global statistical accuracy and extreme hydrometeor representation.

\textbf{Training configuration.} For the Pareto front analysis discussed in the Results section, the loss-weighting tuples, defined as (underprediction penalty, mean-bias weight, intense-event weight, physical hydrometeor weight, adversarial weight), were empirically set to $(1.0, 0.1, 0.0, 0.4, 0.00)$ for the RMSE-oriented setting, $(1.5, 0.2, 0.1, 0.6, 0.01)$ and $(2.0, 0.3, 0.2, 0.8, 0.02)$ for the intermediate Balanced-event settings, and $(3.0, 0.5, 0.4, 1.0, 0.03)$ for the CSI-oriented setting. The final PredHydro-Net configuration follows the selected Balanced-event tuple $(2.0, 0.3, 0.2, 0.8, 0.02)$, and the frequency-domain setting is $\lambda_{\rm spec}/\lambda_{\rm TV}/\lambda_{\rm HF\text{-}reg}=0.50/0.02/0.00$. The seam-consistency weight is fixed at $\lambda_{\rm seam}=0.5$, and the thermodynamic regression weights are $\lambda_{\rm TQ}=3.0$ in standardized space and $\lambda_{\rm phys,TQ}=1.0$ in physical space. The generator is optimized with AdamW with weight decay $10^{-4}$, while the discriminator uses Adam with $\beta=(0.5,0.999)$. The final Pareto fine-tuning uses a two-stage schedule: a short decoder-adaptation phase with the encoder, ST-LSTM, and autoregressive adapter frozen at a generator learning rate of $3\times10^{-5}$, followed by full-model fine-tuning at a generator learning rate of $1\times10^{-5}$ and discriminator learning rate of $2\times10^{-7}$.

\backmatter

\bmhead{Acknowledgements}

This research is supported by the National Natural Science Foundation of China (Grants 42450105 and 41905035), the Science and Technology Innovation Program of Xiongan New Area (2024XAGG0007), and the Science and Technology Development Foundation of Chinese Academy of Meteorological Sciences (Grant 2024KJ007). The authors would also like to acknowledge the 3rd World AI for Science (AI4S) Competition hosted by the Shanghai Academy of AI for Science for motivating this research problem.

\bmhead{Data Availability}
The ERA5 reanalysis data are publicly available from the Copernicus Climate Change Service (C3S) Climate Data Store (CDS) at \url{https://cds.climate.copernicus.eu/datasets/reanalysis-era5-pressure-levels}.\cite{hersbach2020era5} The GFS forecast data are provided by the National Centers for Environmental Prediction (NCEP) and can be accessed via the NOAA National Centers for Environmental Information (NCEI) at \url{https://www.ncei.noaa.gov/products/weather-climate-models/global-forecast}.\cite{han2017updates} The IMERG Final Run precipitation data are available from the NASA Goddard Earth Sciences Data and Information Services Center (GES DISC) at \url{https://disc.gsfc.nasa.gov/datasets/GPM_3IMERGDF_07/summary}.\cite{hou2014global,huffman2020integrated}

\bmhead{Code Availability}
The source code for training and evaluating PredHydro-Net has been deposited in Zenodo: Chen, D. (2026), \textit{Source Code for Physics-Guided Dual Decoding and Spectral Supervision for Global 3D Hydrometeor Prediction}, \url{https://doi.org/10.5281/zenodo.20539888}. Due to large storage requirements, the fully trained model weights will be deposited in a public repository upon acceptance of the manuscript.

\bmhead{Author Contributions}
D.C. designed the model, performed the experiments, analyzed the results, prepared the figures, and wrote the initial manuscript. Y.W. supervised the study, contributed to the research design and interpretation of the results. Both authors reviewed and approved the final manuscript.

\bmhead{Competing Interests}
The authors declare no competing interests.

\bibliography{refs}

\clearpage

\bmhead{Supplementary information}
Supplementary information accompanies this paper.


\setcounter{table}{0}
\renewcommand{\thetable}{S\arabic{table}}
\renewcommand{\theHtable}{S\arabic{table}}

\begin{table}[!htbp]
\centering
\caption{Input channel composition of PredHydro-Net}
\label{tab:input_channels}
\small
\setlength{\tabcolsep}{3pt}
\begin{tabular}{@{}p{0.18\linewidth} p{0.30\linewidth} p{0.36\linewidth} r@{}}
\toprule
Feature group & Input variables & Levels and definitions & Channels \\
\midrule
Basic state variables
& Geopotential ($z$), temperature ($T$), zonal wind ($u$), meridional wind ($v$), specific humidity ($q$)
& 7 pressure levels: 200, 300, 500, 700, 850, 925, and 1000 hPa
& 35 \\
\addlinespace
Hydrometeor input variables
	& CIWC, CLWC, CRWC, and CSWC
& Same 7 pressure levels
& 28 \\
\addlinespace
Physical diagnostic variables
& Relative vorticity ($\zeta$), horizontal divergence ($\nabla \cdot \mathbf{v}$), relative humidity (RH)
& Computed from ERA5 wind, temperature, and humidity fields at the same 7 pressure levels
& 21 \\
\addlinespace
Thickness diagnostics
& Geopotential-thickness differences $\Delta Z$
& Eight selected layer pairs: 925--1000, 850--1000, 850--925, 700--1000, 700--850, 500--1000, 500--850, and 500--700 hPa
& 8 \\
\addlinespace
Geometric coordinate
& Latitude ($\phi$)
& Appended as the 93rd channel
& 1 \\
\midrule
Total
& 92 ERA5-derived meteorological channels + 1 latitude channel
& Model input tensor: $T_{\mathrm{in}} \times 93 \times H \times W$
& 93 \\
\bottomrule
\end{tabular}
\begin{tablenotes}
\item The model uses 92 ERA5-derived meteorological channels plus one appended latitude channel, giving 93 input channels at each input time step.
\end{tablenotes}
\end{table}

\clearpage


\setcounter{figure}{0}
\renewcommand{\thefigure}{S\arabic{figure}}
\renewcommand{\theHfigure}{S\arabic{figure}}

\begin{figure}[!htbp]
\centering
\includegraphics[width=\linewidth]{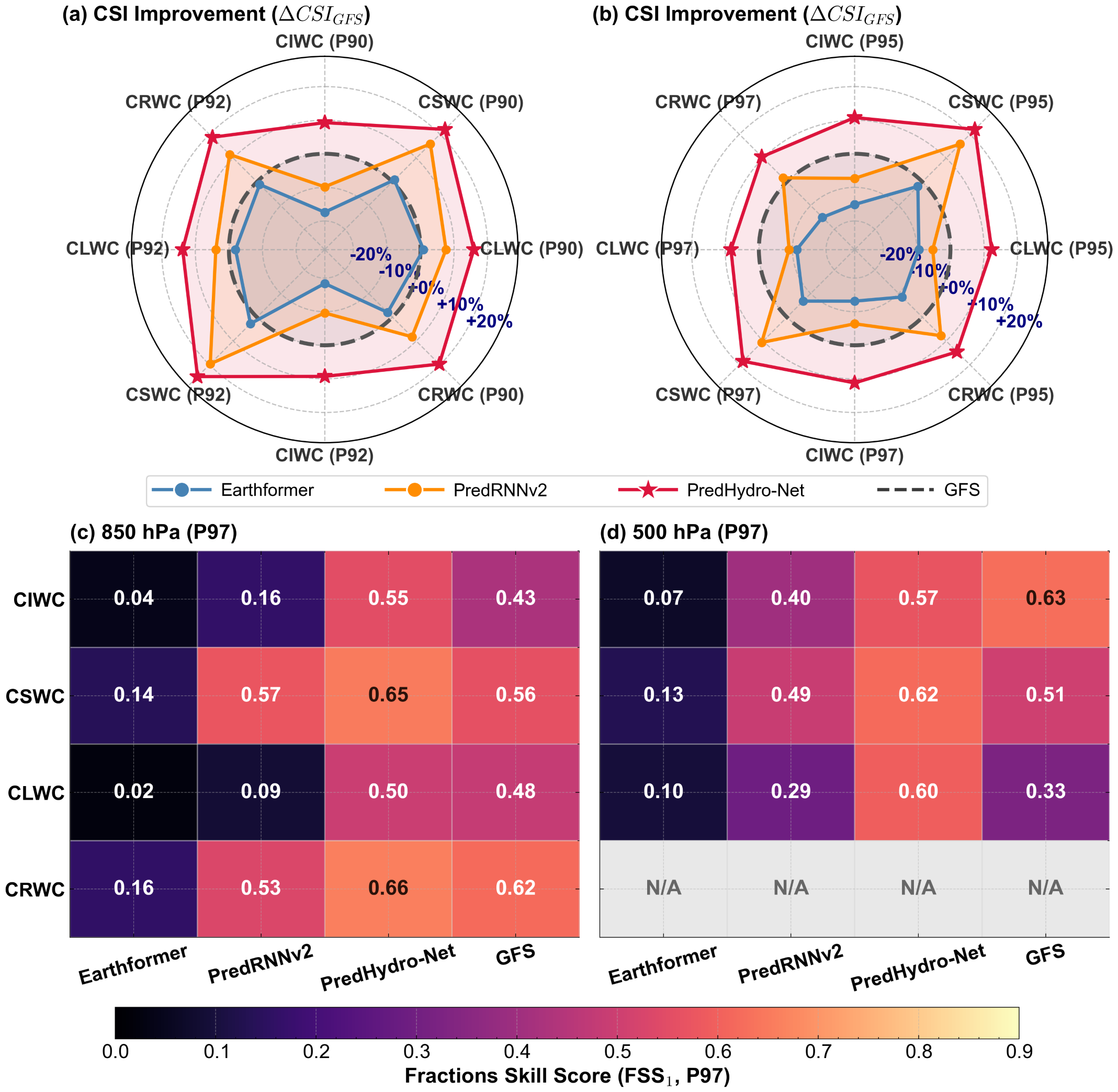}
\caption{Extreme value detection and neighborhood-scale spatial agreement. \textbf{a, b}, Radar charts showing CSI differences ($\Delta CSI_{GFS}$) relative to the GFS baseline across selected hydrometeor species and percentile thresholds. The dashed gray polygon denotes the GFS reference. \textbf{c, d}, Fractions skill score (FSS$_1$) heatmaps at the 97th percentile threshold for 850 hPa and 500 hPa, comparing neighborhood-scale spatial agreement among the AI-based models and the GFS baseline. N/A indicates species--level combinations that are not evaluated because the corresponding ERA5 hydrometeor field is too sparse for a stable percentile-threshold score, such as specific rain water content (CRWC) at 500 hPa.}
\label{fig:supp_extreme_agreement}
\end{figure}

\begin{figure}[!htbp]
\centering
\includegraphics[width=\linewidth]{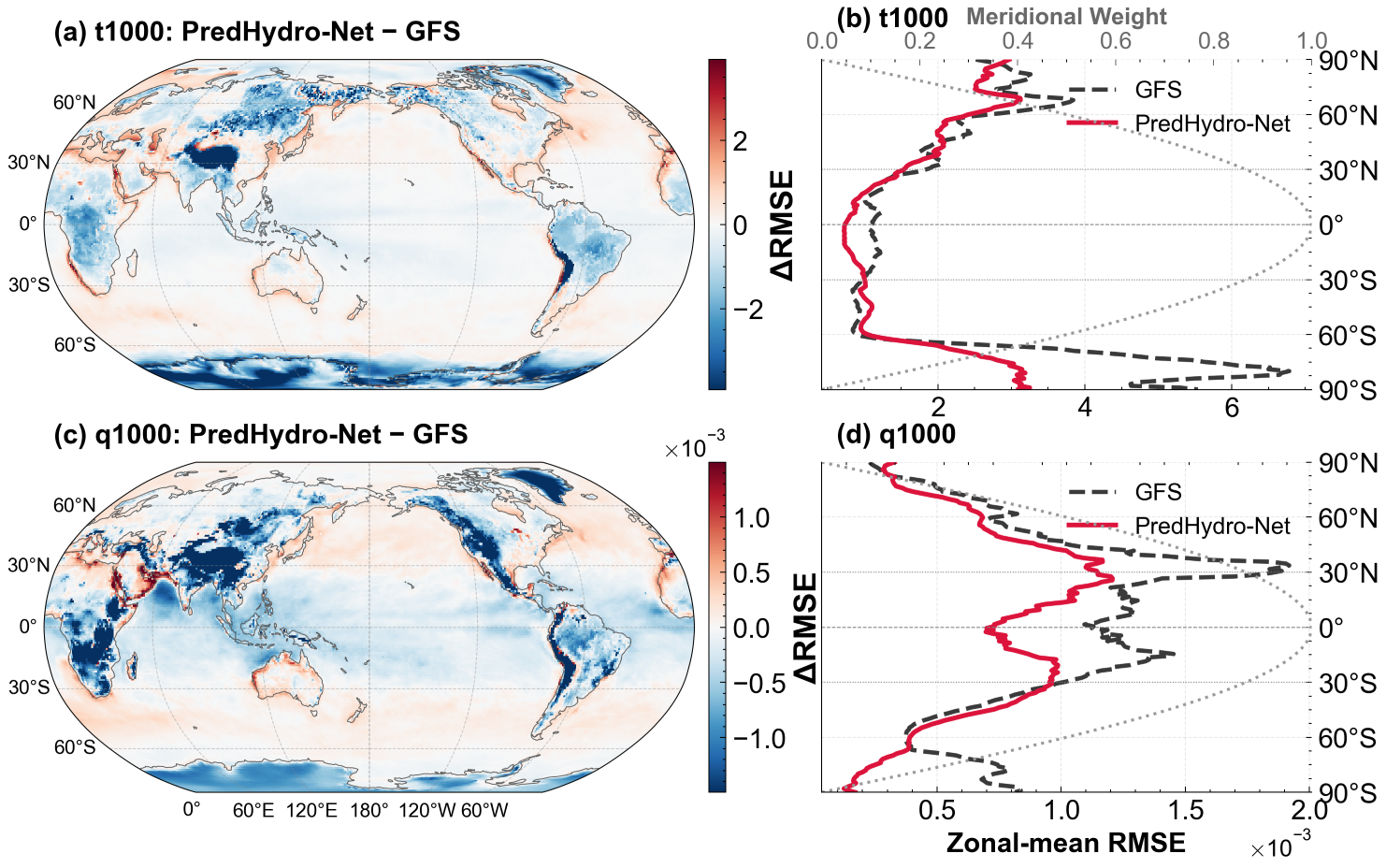}
\caption{Global error patterns for continuous thermodynamic variables. \textbf{a, c}, Spatial maps of RMSE difference ($\Delta$RMSE, PredHydro-Net minus GFS) for temperature (t1000) and specific humidity (q1000) at 1000 hPa. Negative values indicate regions where PredHydro-Net has lower RMSE than GFS under this comparison, whereas positive values indicate lower GFS RMSE. \textbf{b, d}, Corresponding zonal-mean RMSE profiles for GFS and PredHydro-Net, with the dotted curve indicating the meridional weighting used in the summary. Polar regions show localized negative $\Delta$RMSE but receive low meridional weight. These patterns are therefore used to characterize regional error structure rather than to claim a general advantage over GFS for smooth thermodynamic or circulation-related variables.}
\label{fig:supp_tq_errors}
\end{figure}

\begin{figure}[!htbp]
\centering
\includegraphics[width=\linewidth]{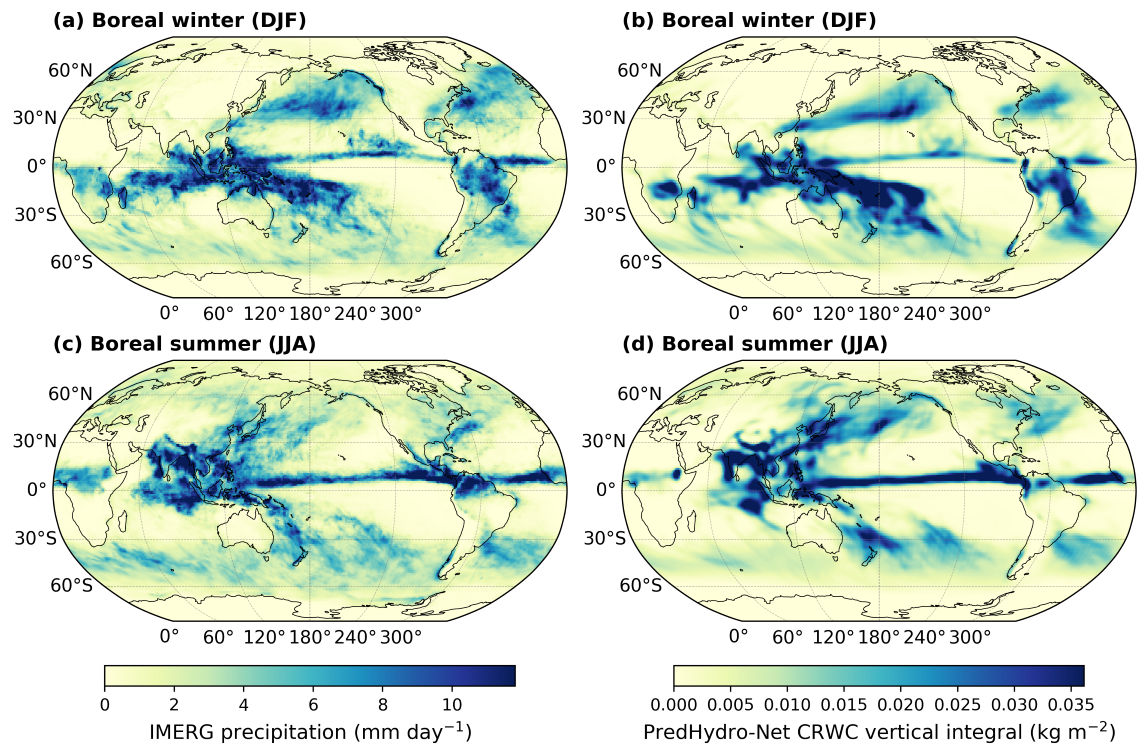}
\caption{Seasonal precipitation and CRWC climatology. \textbf{a, c}, IMERG mean precipitation rate (mm day$^{-1}$) for boreal winter (DJF) and boreal summer (JJA), respectively. \textbf{b, d}, Corresponding PredHydro-Net CRWC vertical integral (kg m$^{-2}$) for the same seasons. This comparison provides a qualitative seasonal check of precipitation-related hydrometeor organization rather than a direct variable-to-variable verification. The seasonal cycle captures the northward migration of the ITCZ and the South/East Asian monsoon systems from DJF to JJA, broadly consistent with the IMERG reference.}
\label{fig:supp_seasonal}
\end{figure}

\end{document}